\documentclass{article}

\usepackage{microtype}
\usepackage{graphicx}
\usepackage{booktabs} 

\usepackage{amsmath}
\usepackage{todonotes}
\presetkeys{todonotes}{inline}{}
\usepackage{amssymb}

\usepackage{subcaption}

\usepackage{colortbl}
\usepackage[draft]{hyperref}

\usepackage[accepted]{icml2018}
\icmltitlerunning{Modular Multi-Objective Deep Reinforcement Learning with Decision Values}

\begin{document}

\title{Modular Multi-Objective Deep Reinforcement Learning with Decision Values}
\author{Tomasz Tajmajer \footnote{Faculty of Mathematics, Informatics and Mechanics, University of Warsaw, Poland; email: \texttt{t.tajmajer@mimuw.edu.pl}}}

\date{}

\twocolumn{}
\maketitle

\begin{abstract}
In this work we present a method for using Deep Q-Networks (DQNs) in multi-objective environments. Deep Q-Networks provide remarkable performance in single objective problems learning from high-level visual state representations. However, in many scenarios (e.g in robotics, games), the agent needs to pursue multiple objectives simultaneously. We propose an architecture in which separate DQNs are used to control the agent's behaviour with respect to particular objectives. In this architecture we introduce \textit{decision values} to improve the scalarization of multiple DQNs into a single action. Our architecture enables the decomposition of the agent's behaviour into controllable and replaceable sub-behaviours learned by distinct modules. Moreover, it allows to change the priorities of particular objectives post-learning, while preserving the overall performance of the agent. To evaluate our solution we used a game-like simulator in which an agent - provided with high-level visual input - pursues multiple objectives in a 2D world. 

\end{abstract}

\section{Introduction}

Many recent works on Reinforcement Learning focus on single-objective methods such as Deep Q-learning \cite{mnih-atari-2013,Mnih2015}. As those methods provide great performance in task such as playing video games, many real-life problems require satisfying multiple objectives simultaneously.
In single objective reinforcement learning the agent receives a single reward each time it performs an action. In multi-objective reinforcement learning (MORL) the agent receives multiple rewards - one for each objective. In particular, agents dealing with complex environments, such as autonomous robots or agents playing real-time video games, need to pursue multiple, often conflicting objectives. 

  To have a graspable example, lets consider an autonomous cleaning robot, which is able to clean floors, navigate through obstacles and autonomously return to charging station. The observable aggregated behaviour of such robot may be decomposed into three sub-behaviours: collision avoidance (ca), floor cleaning (fc) and recharging (rg). We may describe the objectives of the robot for each identified sub-behaviour in a multi-objective manner, or we can aggregate the sub-behaviours and define a single objective. In the former case, the robot-agent will receive a set of three rewards ([$r_{ca}$, $r_{fc}$, $r_{rg}$]) after each action. If the robot collides with a wall, it receives a negative reward related to collision avoidance ($r_{ca}$), yet the rewards related to floor cleaning and recharging do not depend on this event. However, in single-objective case, the robot will receive only one reward value ([$r$]) dependent on any of the three sub-behaviours. In case of collision, the the single-objective robot will receive a negative reward, but it will be indistinguishable from any negative reward provided with respect to other sub-behaviours such as depletion of batteries.

In single objective scenarios, we may find an optimal policy for which the sum of rewards collected by the agent is the highest possible. Methods such as Q-learning should converge to optimal policies \cite{Sutton_RL}. However, for multi-objective problems, many such optimal policies may exist, depending on the trade-offs between satisfying particular objectives \cite{MORL_overview}.

Autonomous agents, such as our example cleaning robot, are not really independent - they usually have a purpose defined by another agent: human. This aspect is often neglected in the literature, but is significant when considering practical applications of intelligent agents in robotics, automation or even when designing AIs for video games (always winning AI is not the one that many humans would like to play against). Our cleaning robot may follow a policy for which collision avoidance has greater importance than floor cleaning - in such case the robot should focus on avoiding collisions even at the cost of worse performance at floor cleaning. It is however for the user of such robot to decide, what should be the proportion between carefulness and cleanliness. The user may even want to fully disable some functions (behaviours) of the robot. Yet, state of the art reinforcement learning methods, such as Deep Q-Learning, do not allow to modify the behaviour of the agent after it was trained.

We see that when considering practical applications it is desired to have a multi-objective reinforcement learning method with the following features available post-learning: 1) ability to select the sub-set of pursued objectives and 2) ability to change the impact of particular objectives on the overall policy of the agent. 
As we will show later, the method presented in this paper posses those features.

Multi-objective problems may be approached using \textit{single-policy} or \textit{multi-policy} methods. The simplest single-policy method uses a \textit{scalarization function} \cite{scalarization}, which converts multiple objectives into a single objective. Scalarization methods utilize a weight matrix to obtain a single score from multiple action-value functions. Some techniques assign linear priorities to objectives \cite{optimal_policies_with_multiple_criteria, Vamplew2011}. This allows to obtain a single optimal policy with respect to objectives ordered by those priorities.

In contrast to single-policy methods, multi-policy MORL methods are used for find a set of policies. Their aim is to approximate the Pareto front of policies~\cite{MORL_overview}. In multi-policy methods, the preference of objectives does not need to be set a priori as a Pareto optimal policy for any preference may be obtained at runtime~\cite{MORL_pareto_dominating_p}.

A natural approach in MORL is to use separate learning modules for each objective~\cite{modular_sarsa}. Modularity allows to decompose the problem into components that are to some extent independent \cite{parallel_rl_multiple_rewards}; modularity may be required for providing features desired in practical applications that were listed earlier. Some works deal with transforming complex single-objective problems to many simpler objectives~\cite{multiobjectivization}. Such methods may be used to benefit from modular approach while solving single-objective problems.

Although Deep Q-Networks gained much attention in recent years, not many works consider the use of DQNs in multi-objective problems. Recently authors of \cite{mo_drl} proposed a multi-policy learning framework that utilizes Deep Q-Networks.

Learning behaviours in embodied agents, such as robots, is a problem well fitted for reinforcement learning methods. In \textit{embodied artificial intelligence}, the idea of \textit{parallel, loosely coupled processes} \cite{understanding_intelligence} is proposed as a principle for designing embodied agents. It states, that the control logic for embodied agents should consists of many independent components dedicated for particular aspects of the agent's behaviour. The aggregated behaviour of an agent emerges from cooperation or competence among those components. 

In this work we will present a method for combining multiple Deep Q-Networks for solving multi-objective problems. We will introduce decision values used for more advanced scalarization of multiple Q-functions. Furthermore we will combine decision values with user define priorities, to have an architecture that can dynamically adapt its behaviour with respect to user's preferences.

In section \ref{background} we will briefly describe single- and multi- objective reinforcement learning. Next, in section \ref{main_section} we will describe how many separate DQNs may be used together and we will define decision values. In section \ref{eval} we will present a simple 2D game - a virtual environment including an autonomous agent that has a local (situated) sensory inputs and may pursue different objectives. Finally in the last section we will evaluate our solution and present the results of our experiments.

\section{Background}
\label{background}
\subsection{Single Objective Reinforcement Learning}

In the single-objective reinforcement learning an agent interacts with the environment by perceiving the state  $s_t \in S$ and performing an action $a_t \in A$ for each step $t$. The actions are chosen by the agent according to some policy $\pi$. After performing an action, the agent receives a reward $r_t$. Then the agent observes the next state $s_{t+1}$ and the process repeats. The goal of the agent is to maximize the expected discounted reward $R_t = \sum^{\infty}_{k=0}{\gamma^k r_{t+k}}$, where $\gamma \in [0,1]$ is the discount factor.

In Q-learning actions are selected based on $Q(s,a)$, which represents the expected discounted reward for performing action $a$ in state $s$. For given state s, $a_t = \underset{a}{\arg\max} Q(s, a)$ is the optimal action. The policy of an agent, denoted by $\pi$, is the probability of selecting action $a$ in state $s$. If the agent always selects the optimal action, then we say that it follows an optimal policy $\pi_\star$. Knowing the $Q(s,a)$ allows to create an optimal policy simply by selecting the action with the highest Q-value. Deep Q-learning utilizes Deep Neural Networks for approximating $Q(s,a)$ values, thus enabling this method to be used in many real-world applications. Deep Q-Networks \cite{Mnih2015} may be used used with high-level visual inputs such as those provided by video games.

\subsection{Multi-Objective Reinforcement Learning}
\label{morl_section}

We may consider a more complex reinforcement learning scenario in which multiple objectives are pursued by the agent. Let $O$ be the set of objectives of an agent.  
We may assign a priority $p$ to each objective $o \in O$ such that $o_k$ will have lower priority than $o_j$ when $p(o_k) < p(o_j)$. For further analysis we will assume that $\forall_{o \in O} p(o) \geq 0$, so that priorities may be interpreted as weights.

The agent, instead of a single reward, receives a vector of rewards at each time-step $t$ with respect to each objective $o_i$, i.e: $\vec{r}_t = \left[r_{1,t}, r_{2,t}, \dots \, r_{n,t} \right]$, where $r_{i,t}$ corresponds to objective $o_i$. For each objective $o_i$ and step $t$ we may define the discounted return as:
\begin{equation}
R_{i,t} = \sum^{\infty}_{k=0}{\gamma^k r_{i,t+k}} 
\end{equation}
Moreover, for each objective $o_i$ there is a Q-function $Q_i(s,a)$ that represents the expected discounted return $R_{i,t}$, i.e: $Q_i(s,a) = \mathbb{E}\left[ R_{i,t} \, | \, s_t = s, a_t = a\right]$. 
We may define a vector of Q-functions, which includes $Q(s,a)$ for each objective $o_i$:
\begin{equation}
\vec{Q}(s,a) = \left[Q_1(s,a), Q_2(s,a), ..., Q_n(s,a)\right]
\end{equation}
The function $Q_i(s,a)$ may be used by the agent to determine the optimal action with respect to objective $o_i$ at time-step $t$, given state $s_t$:
\begin{equation} 
\label{argmax_Q}
a_{i,t} = \underset{a}{\arg\max}\,Q_i(s_t, a)
\end{equation}

The vector $\vec{a}_t = \left[a_{1,t}, a_{2,t}, ..., a_{n,t}\right]$ consists of actions optimal with respect to particular objectives at a given time-step $t$. Because at each step, the agent may perform only a single action, a method of reducing $\vec{a}_t$ to a single action is required. 

A common method for selecting a single action is the scalarization \cite{scalarization} of $\vec{Q}(s,a)$ using some scalarization function and a weight vector $\vec{w}$. Typically a linear scalarization is applied, so that:
\begin{equation}
SQ(s,a) = \sum^N_{i=1} {w_i Q_i(s,a)}
\end{equation}
Then $SQ(s,a)$ may be used as in equation \ref{argmax_Q} to select an action. The weight wector in this case corresponds to priorities assigned to particular objectives.

In the further sections of this paper, we will show how to apply scalarization in Deep Q-Networks and we will introduce Decision Values to dynamically adjust the weights for improved performance of the agent. For simplicity, further in the text we will use the index $i$ to note that a particular value or function is defined for any objective $o_i$, and by $N$ we will define the number of objectives.

\section{Using multiple DQNs}

\label{main_section}

We have considered an agent that have multiple objectives, receives rewards with respect to those objectives and has a separate Q-function for each objective. In this section we will describe how to merge q-values obtained from Deep Q-Networks for different objectives and how the impact of particular DQNs on the behaviours of the agent may be controlled by using Decison Values. Finally we will describe the learning process utilizing DQNs with Decision Values. We will refer to our method as to Multi-Objective Deep Q-Network with Decision Values (MODQN-DV).

\subsection{Combining Q-values}

\label{combining_dqns}

In case of multi-objective agent, we may use a separate DQN as an approximator for each $Q_i(s,a)$ in the $\vec{Q}(s,a)$ vector. Such agent would be controlled by multiple Deep Q-Networks working in parallel. Each DQN provides a list of q-values and we want to use q-values from all DQNs to select a single action $a$ that will be performed by the agent

Let us define a vector $\vec{q}_i$ that consists of q-values provided by $Q_i(s,a)$ for each possible action $a \in A$ and a single objective $o_i$, i.e.:
\begin{equation}
\vec{q}_i = \left[Q_i(s, a_0), Q_i(s, a_1), ..., Q_i(s, a_j)\right]
\end{equation}
In the single-objective case the optimal action $a$ would be equal to $a_j$ for such $j$ that $\vec{q}_{i,j} = \max \vec{q}_i$. For multi-objective case we can use scalarization to sum up all $\vec{q}$ vectors and then select the action corresponding to the maximal value of such scaled q-value vector. In this approach, q-values may be interpreted as votes of certain DQN, which are summed-up and the highest-voted action is selected. We need to stress here that simply adding the vectors does not produce a meaningful result yet. The q-values produced by different Q-functions are not scaled. In general q-values may be any real numbers. If we want them to represent votes for particular actions, each $\vec{q}_i$ vector needs to be rescaled to $[0,1]\subseteq\mathbb{R}$. Many approaches for scaling the vector may be applied. In our experiments we use the following scaling function for which $\min(\vec{q}_i)$ is mapped to $0$ and $\max(\vec{q}_i)$ to $1$:

\begin{equation}
 scale(\vec{x}) = \frac{\vec{x} - \min(\vec{x})}{\max (\vec{x} - \min(\vec{x}))}
\end{equation}

The scalarized q-vector is then defined as:
\begin{equation}
 \vec{q_s} = \sum_{i=1}^N w_i scale(\vec{q_i})
\end{equation}

Now, using the rescaled $\vec{q}_i$ vectors  we can sum them up and select one action with the highest total q-value. For example, let have actions ${a_1, a_2, a_3}$, weight vector $\vec{w} = [1, 1, 1]$, objectives $o_1$, $o_2$ and corresponding q-vectors $ \vec{q}_1 = [0, 0.6, 1]$ and $ \vec{q}_2 = [1, 0.5, 0]$. Adding them will result in vector $[1, 1.1, 1]$, for which the second element is the maximal, thus the corresponding action $a_2$ should be selected.

\subsection{Decision Value}

\label{decision_value}

\begin{figure}[!t]
\centering
\includegraphics[width=4.5in, trim={0.7in, 1.5in 0 0}, clip]{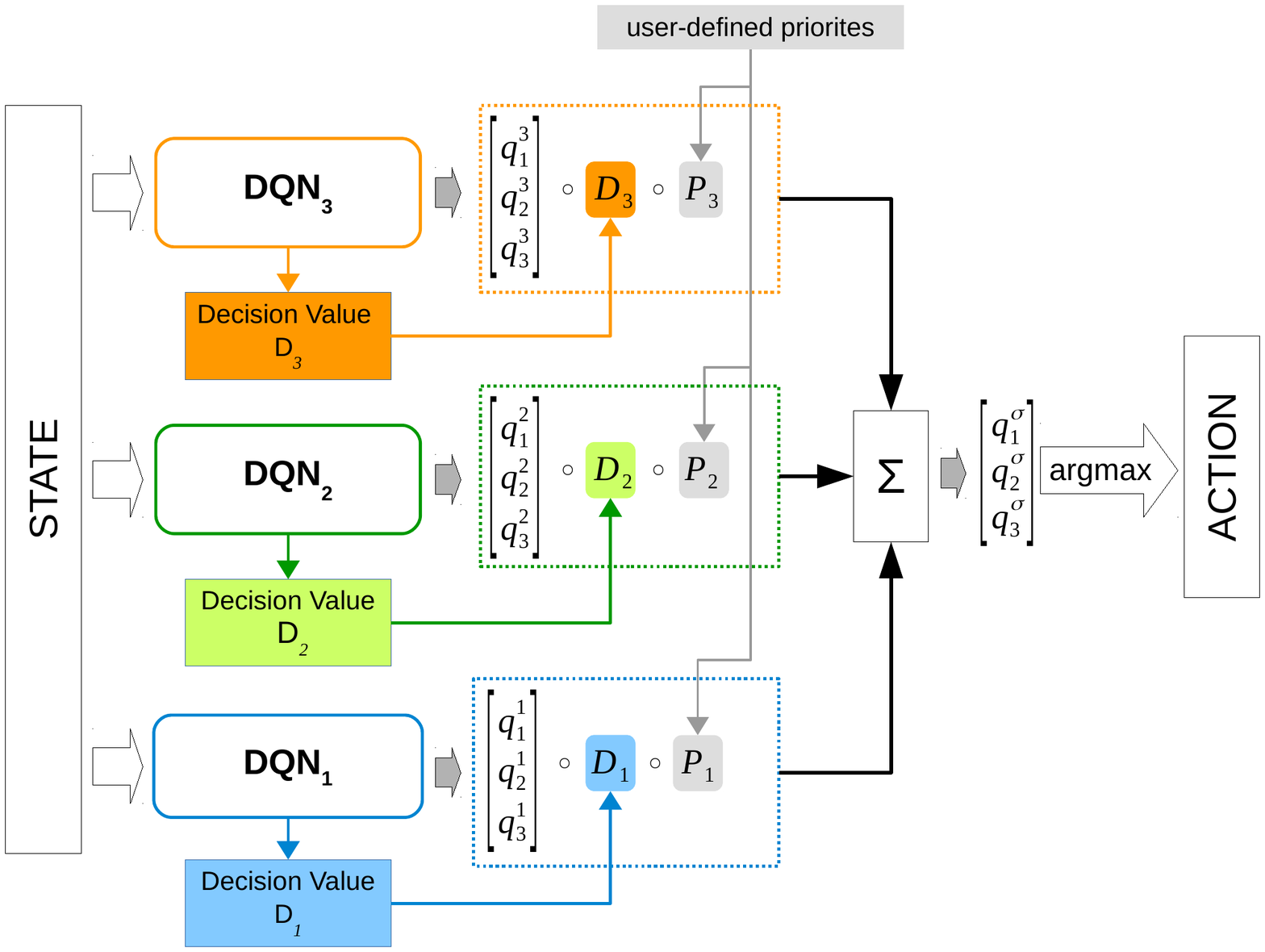}
\caption{Three Deep Q-Networks are working in parallel based on the same sensory input. Each DQN corresponds to different task pursued by the agent. Each DQN has an additional decision value output which acts as a dynamic weight used while summing up q-vectors from particular DQNs. User defined priorities are also used for weighting the decision from particual DQNs.}
\label{3layer}
\end{figure}

The scalarization allows to combine outputs from multiple DQNs. However, such a combination does not guarantee a meaningful action selection. Let us return to previous to examples and consider a vacuum cleaner approaching a wall; actions ${a_1, a_2, a_3}$ correspond to turning left, going straight, and turning right respectively. If the vacuum cleaner perform the action proposed in $q_1$ it will turn right, alternatively if it uses $q_2$ then it will turn left. Using the sum will however lead to going straight forward and hitting the wall. So while both DQNs suggested a meaningful action, their sum is not meaningful at all. We see that using constant weights while summing q-values does not provide a solution for this problem.  

To solve this issue, we would need to dynamically choose which q-value vectors are more important in a particular state. In other words, we would like to have  a meta-policy for choosing the actual policy of the agent. However, as the agent pursues many objectives, it is hard to define this meta-policy with respect to all objectives. To overcome this problem we propose to indicate the \textit{value} of the \textit{decision} provided by each DQN with respect to corresponding objective pursued by the agent. 

The proposed \textit{decision values} may be indicated independently by each DQN based on the current state and used as additional weights while summing up q-value vectors. Going back to the previous example: let assume that $q_1$ is the output from DQN associated with collision avoidance and $q_2$ is the output from DQN associated with cleaning. As the robot approaches a wall, the decision regarding collision avoidance is clearly more important than the decision regarding cleaning. This is because if the robot does not make any decision, it will collide with the wall and receive a negative reward with respect to collision avoidance objective. However, not making the decision will not affect cleaning objective (assuming that the cleanliness of the floor in front of him is not different than in other places). Thus, at this particular state the value of $q_1$ is higher than the value of $q_2$ and $q_1$ should be summed with a higher weight.

We may define the decision value signal $d \in [0,1]\subseteq\mathbb{R}$, and by $d_i$ denote the decision value associated with $DQN_i$. Now the
scalarized q-vector would use decision values instead of constant weights:
\begin{equation}
 \vec{q_d} = \sum_{i=1}^N d_i scale(\vec{q_i})
\end{equation}

We may additionally include the external preferences indicated by values of priorities $p_i$ assigned to objectives as introduced in \ref{morl_section}. This way the q-values will be scaled both by dynamic decision values and static priorities. Moreover, for technical reasons, we need to add $\vec{\mu}$, which is a vector containing very small random values. This will ensure that in a rare cases when all decision values are equal to $0$, a random action will be chosen. Finally the scaled, decision value- and priority- weighted q-value vector denoted by $\vec{q_\sigma}$ is equal to:

\begin{equation}
 \vec{q_\sigma} = \vec{\mu} + \sum_{i=1}^N d_i p_i scale(\vec{q_i})
\end{equation}

\subsection{Acquiring values of decisions}

Now, as we have a method of applying decision values in the scalarization of multiple objectives, let us explain in more details how decision values are defined and how they can be learned by reinforcement learning. 

First we should consider how objectives of an agent are defined. Again let us refer to the vacuum cleaning robot example. If the agent had only two objectives: a) to seek dirt and b) to avoid colliding with obstacles, then we could define two reward/terminal states:  state A - state in which dirt is collected, state B - state in which the robot is colliding with something. There is a notable difference between those two states. In the first case, the agent should be rewarded positively, but in the the latter case, it should be rewarded negatively. Moreover, if the agent is not in any of those states, it should be not rewarded at all. We can describe the first objective as being \textit{attractive} (as it attracts the agent by positive rewards) and the second as being \textit{repulsive} (as it repulses the agent by negative rewards). Many problems in robotics, games or other fields of AI may be presented using a set of attractive or repulsive objectives. In particular some problems may be decomposed into such set of objectives to promote more granular learning and control. Such decomposition is usually simpler and more intuitive compared to more advanced reward shaping techniques.

Let us consider an agent moving in a state-space with attractive and repulsive states. As the agent approaches one of those states, it becomes more critical to perform an action that will either move the agent towards such state or away from it. The value of the decision made with respect to an objective near a rewarding state rises as the distance to this state becomes shorter. This is a simple and intuitive heuristic: if an agent pursues multiple equally weighted objectives, then it probably should focus most on the objective that is already very close to being accomplished.

We can thus create a \textit{decision reward} - the reward provided to the agent for performing a decision - which would be simply the absolute value of the reward provided with respect to an objective: $\rho_i = abs(r_i)$. Now we can define the decision value as a \textit{state-value} function \cite{Sutton_RL}, returning the value of the state $s$ under policy $\pi$, with respect to the decision rewards of a particular objective:

\begin{equation}
D_i(s) = \mathbb{E}_{\pi}\left[ \sum_{k=0}^{\infty} \gamma^k\,\rho_{i,t+k+1} \,\middle|\, s_t = s \right]
\end{equation}

Such defined decision value will provide high values around rewarding states (either positive or negative) and low values in states which are far from rewarding states. In any state, the decision value will provide the importance of particular objective. The proposed decision value function will hence provide values representing the chances of achieving a rewarding state (with respect to some objective $o$) given the current state $s$ and following policy $\pi$. Where policy $\pi$ is the policy provided by the Q-function for a particular objective.  

It is important to note, that the decision value, as defined, can not be directly used for scalarization, because its value may be any positive number. Moreover, the range of the values provided for different objectives may be very broad. To overcome this problem, the decision value needs to be scaled to be in range $[0, 1]$ as noted in section \ref{decision_value}. However, the unscaled decision value is needed during learning as it will be shown in the next section. We will therefore denote the unscaled decision value by $D_i$ and define the scaled decision value by $d_i$ as follows:
\begin{equation}
d_i = \sigma\left (\frac{(D_i - \alpha_i)} {\beta_i} \right )
\end{equation}

Where $\sigma$ is the sigmoid function; $\alpha_i$ and $\beta_i$ are derived during learning: $\alpha_i$ is an approximation of the mean value of $D_i$, while $\beta_i$ is an approximation of the standard deviation of $D_i$.

\subsection{Learning}
\label{learning}

Have defined decision values, we may move to the method of learning such values along with learning policies for particular objectives. We use Deep Q-Networks to approximate the values of Q-functions. Following the state of the art in this field a $DQN$ provides the approximated function $Q(s,a; \theta)$, where $\theta$ are the learnable parameters of the neural network. As in our model we use multiple DQNs, there is a function $Q_i(s, a; \theta_i)$ for a $DQN_i$ related to objective $o_i$. Each $DQN_i$ is optimised iteratively, using the following loss function for each iteration $j$:
\begin{equation}
\begin{split} 
L^{Q}_{i,j}(\theta_{i,j}) &= \mathbb{E}_{(s,a,r_i,s')\sim U(M_i)} [ ( r_i + \\
&+ \gamma \underset{a'}{\max}\,Q_i(s',a'; \theta_{i,j}^-) - Q_i(s,a; \theta_{i,j}) )^2 ]
\end{split} 
\end{equation}
As introduced in \cite{mnih-atari-2013}, there are in fact two neural networks involved in the learning process of a single DQN. The \textit{on-line network} $Q_i(s, a; \theta)$ is updated at each iteration, while the \textit{target network} $Q_i(s', a';\theta^-)$ is updated only each $K$ iterations. Moreover \textit{experience replay} is used to further improve the learning process. The agent stores experienced states, actions and rewards in a \textit{replay memory} $M_i$ for each $DQN_i$ respectively. Then at each iteration, each $DQN_i$ is trained using a sample of past experiences selected uniformly at random from the corresponding replay memory $M_i$. Those samples are used as mini-batches for gradient descent optimization. 

The Decision Value may be updated using TD-learning \cite{Sutton_RL} similarly as for any state value function, by using the following update rule:
\begin{equation}
D_i(s_t) \leftarrow D_i(s_t) + \alpha \left[\rho_i + \gamma D_i(s_{t+1}) - D_i(s_t)\right]
\end{equation}
As we use a neural network for approximating $D_i(s)$, we may define the loss function as follows:
\begin{equation}
\begin{split}
L^{D}_{i,j}(\theta_{i,j}) &= \mathbb{E}_{(s, \rho_i, s')\sim U(M_i)} [ ( \rho_i +\\
&+ \gamma D_i(s'; \theta_{i,j}^-) - D_i(s; \theta_{i,j}) )^2 ]
\end{split}
\end{equation}
The decision value is provided by an additional output of the DQN and the learning procedure is analogical to Q-function. Moreover the decision value requires scaling, for which the parameters $\alpha$ and $\beta$ need to be learned. If we include $\alpha$ and $\beta$ in the neural network parameters $\theta$, then the additional loss function for the decision value scaling would be defined as:
\begin{equation}
\begin{split}
L^{d}_{i,j}(\theta_{i,j}) &= \mathbb{E}_{(s)\sim U(M_i)} [(0.5 - \sigma(D_i(s; \theta_{i,j})))^2 + \\
&+ (1 - \max_s(D_i(s; \theta_{i,j})) - \min_s(D_i(s; \theta_{i,j})))^2]
\end{split}
\end{equation}
The neural network is optimized using a combined loss function for Q-values, decision values and scaling of the decision values:
\begin{equation}
\begin{split} 
L_{i,j}(\theta_{i,j}) &= L^{Q}_{i,j}(\theta_{i,j}) +  L^{D}_{i,j}(\theta_{i,j}) + L^{d}_{i,j}(\theta_{i,j})
\end{split} 
\end{equation}

\section{Evaluation}
\label{eval}
\subsection{Cleaner - a 2D game-like virtual environment}

\begin{figure}[bt]
\centering
\includegraphics[width=4.5in, trim={1.45in, 2.2in, 2in, 2in}, clip]{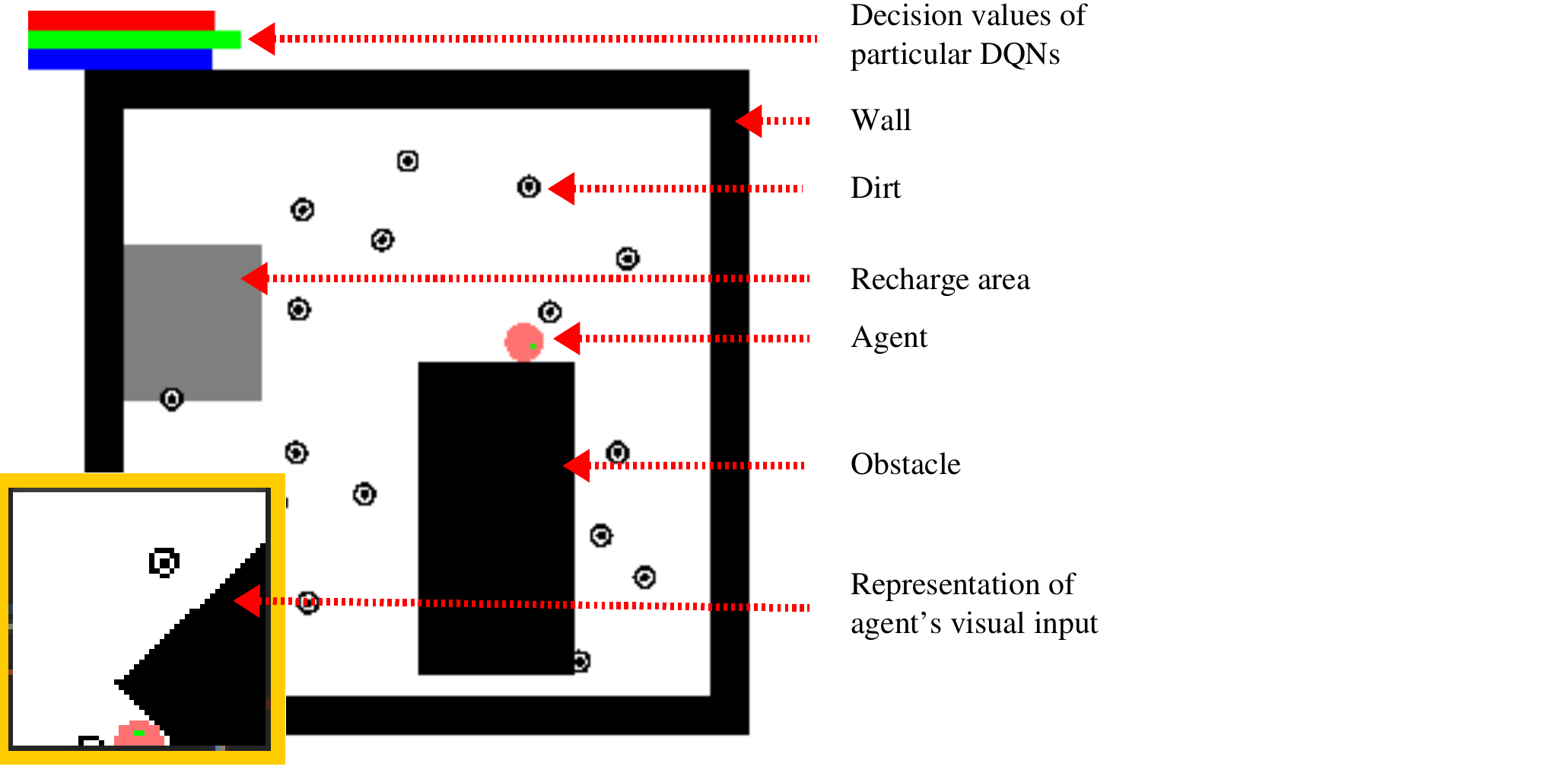}
\caption{Cleaner - a game-like virtual environment with agent pursuing multiple-objectives. The environment consists of the agent, walls, obstacles, recharge area and dirt. The agent perceives the environment by a visual input (a view from the top limited to a square located in the front of the agent). Agent may move forward and turn; its area of movement is limited by walls. Agent has three objectives: avoid walls, consume dirt and recharge.}
\label{game}
\end{figure}

To evaluate the solution presented in this paper we created \textit{Cleaner} - a simple game-like virtual environment, simulating the behaviour of an autonomous vacuum cleaner. The environment consists of an agent, walls, recharge areas and dirt consumable by the agent. Cleaner is presented in Figure \ref{game}. The agent is a circular object that may move around the map by performing one of three actions: move forward, turn left and turn right. The map is a continuous space. The agent perceives the environment only by visual sense, i.e. a $W$ x $H$ pixel (width and height) rectangle situated in front of him. This visual input is converted to gray-scale (8bit). Agent's world (white) is surrounded by walls and filled with obstacles (black rectangles) which agent can not pass. Agent may pick up dirt and recharge itself. Dirt is indicated by three small coaxial circles (black), while recharging field is indicated by a gray rectangle. Dirt re-spawns at random positions on the map after being consumed by the agent. The quantity of dirt, recharge fields and obstacles is constant during the episode. Eater is a simplified simulation of a mobile robot moving on a flat surface (e.g. floor) with a video camera attached at the top of the robot pointed towards the floor.  

The agent has a battery level $E \leq E_{max}$, which is decreased at each time step by $E_{step}$. The battery level may be increased when the agent enters the recharging area by $(1 - E) \cdot 0.1$ each step. An episode ends when the agent's energy level drops to $0$ or when $2000$ steps pass. The agent starts each game with initial battery level $E = E_{start}$. The position of dirt, recharge fields and obstacles as well as the initial position of the agent are chosen randomly at the start of the episode.

The agent has three objectives: (ca) collision avoidance, (fc) cleaning and (rg) recharging.

The rewards for particular objectives are as follows: objective (ca): $-1$ for collision, $0$ otherwise; objective (fc): $+1$ for collecting dirt, $0$ otherwise; objective (rg): $-1$ for for each step when $E < 0.1$, $(1 - E) \cdot 0.1$ while charging and $0$ otherwise. 

In all experiments described in this chapter, the game options were as follows: $E_{start} = E_{max} = 1.0$, $E_{step} = 0.001$. The size of the agent sight rectangle is $W = 50$\,px, $H = 50$\,px. The quantity of food is 20. The numer of obstacles varies randomly from 1 to 5, and the number of charging areas varies randomly from 1 to 3.

\subsection{MODQN-DV implementation} 
\label{implementation}

\begin{table}[bt]
\centering
\caption{MODQN-DV learning hyperparameters}
\label{dqn_params}
\begin{tabular}{ll}
  \hline\noalign{\smallskip}
  Parameter & Value \\
  \hline\noalign{\smallskip}
  learning steps & 1000000 \\
  replay memory size & $10000$  \\
  target network update rate\,\,\,\, & $1000$  \\
  learning rate & $0.001$  \\
  $\epsilon$ start value & $1$  \\
  $\epsilon$ end value & $0.1$  \\
  $\epsilon$ end step & 100000  \\
  discount & $0.99$ \\
  batch size & $32$ \\
  optimizer & Adam  \\
  \hline\noalign{\smallskip}
\end{tabular}
\end{table}

Our implementation of the MODQN-DV was based on the baseline DQN implementation \cite{baselines} developed by OpenAI using TensorFlow\cite{tensorflow2015-whitepaper}. We expanded the standard DQN with additional decision value outputs and mechanism for scalarizing q-values from multiple DQNs. Each single DQN in a MODQN-DV consist of a convolution network with three convolution layers and no pooling layers, followed by a fully connected layer and the output layer. Dueling \cite{dueling} and double q-learning \cite{double_q_learning} were used. The additional decision value output is a single neuron linear layer connected to the state score layer used for dueling. 

The parameters of the convolution network were kept default as provided in the baselines implementation. The size of the fully connected layer in our models is set to 128, and the size of the input image is our case is 50x50x1, thus the q-values are provided based only on an image input from a single state. The memory replay was modified to store rewards with respect to all objectives separately. The prioritized experience replay\cite{per} was not used in our implementation. The hyperparameters used for training DQNs during evaluation are presented in table~\ref{dqn_params}.

During training of the MODQN-DV, loss functions are used as specified in section \ref{learning}. DQNs for all objectives are trained simultaneously and scaled decision values are used for scalarization during learning. 

\subsection{Experiments}

\begin{table*}[tb]
    \caption{Evaluation results}
    \begin{subtable}{.5\linewidth}
    \caption{with decision values enabled}
    \label{eval_with_dv}
    \begin{center}
    \begin{small}
    \begin{sc}
    \begin{tabular}{p{0.15cm} p{0.15cm} p{0.15cm} l l l ll}
    \toprule
    $p_{ca}$ & $p_{fc}$ & $p_{rg}$ & $\Sigma r_{ca}$ & $\Sigma r_{fc}$ & $\Sigma r_{rg}$ & $\Sigma\Sigma r_i$ \\
    \midrule
    1            & 1            & 1            & -88.4           & 47.6           & -35.0           & -75.9     \\
    $\Delta_{baseline}$&              &              & ---             & ---            & ---             & ---       \\
    \midrule
    \textbf{1}   & 0            & 0            & \textbf{-51.9}  & 24.0           & -46.2           & -74.1     \\
    $\Delta_{baseline}$&              &              & \cellcolor{green!25}\textbf{41.4\%} & \cellcolor{red!25}-49.6\%        & \cellcolor{green!25}-32.0\%         & \cellcolor{green!25}2.37\%    \\
    \midrule
    0            & \textbf{1}   & 0            & -303.0          & \textbf{50.0}  & -40.3           & -293.3    \\
    $\Delta_{baseline}$&              &              & \cellcolor{green!25}-242.7\%        & \cellcolor{green!25}\textbf{5.1\%} & \cellcolor{green!25}-15.2\%         & \cellcolor{green!25}-286.74\% \\
    \midrule
    0            & 0            & \textbf{1}   & -311.8          & 20.6           & \textbf{-35.9}  & -327.2    \\
    $\Delta_{baseline}$&              &              & \cellcolor{red!25}-252.6\%        & \cellcolor{red!25}-56.7\%        & \cellcolor{red!25}\textbf{-2.7\%} & \cellcolor{red!25}-331.32\% \\
    \midrule
    \textbf{0.5} & 0.3          & 0.2          & \textbf{-45.7}  & 42.9           & -39.2           & -42.1     \\
    $\Delta_{baseline}$&               &              & \cellcolor{green!25}\textbf{48.4\%} & \cellcolor{red!25}-9.9\%         & \cellcolor{green!25}-12.2\%         & \cellcolor{green!25}44.55\%   \\
    \midrule
    \textbf{0.5} & 0.2          & 0.3          & \textbf{-68.4}  & 38.3           & -39.5           & -69.6     \\
    $\Delta_{baseline}$&               &              & \cellcolor{red!25}\textbf{22.7\%} & \cellcolor{red!25}-19.5\%        & \cellcolor{green!25}-12.9\%         & \cellcolor{red!25}8.27\%    \\
    \midrule
    0.2          & \textbf{0.5} & 0.3          & -143.7          & \textbf{51.3}  & -33.2           & -125.6    \\
    $\Delta_{baseline}$&               &              & \cellcolor{green!25}-62.5\%         & \cellcolor{green!25}\textbf{7.9\%} & \cellcolor{green!25}4.9\%           & \cellcolor{green!25}-65.63\%  \\
    \midrule
    0.3          & \textbf{0.5} & 0.2          & -90.0           & \textbf{50.2}  & -34.7           & -74.4     \\
    $\Delta_{baseline}$&               &              & \cellcolor{green!25}-1.7\%          & \cellcolor{green!25}\textbf{5.6\%} & \cellcolor{green!25}0.9\%           & \cellcolor{green!25}1.93\%    \\
    \midrule
    0.2          & 0.3          & \textbf{0.5} & -140.6          & 45.2           & \textbf{-34.7}  & -130.1    \\
    $\Delta_{baseline}$&               &              & \cellcolor{red!25}-59.0\%         & \cellcolor{green!25}-4.9\%         & \cellcolor{green!25}\textbf{0.6\%}  & \cellcolor{green!25}-71.54\%  \\
    \midrule
    0.3          & 0.2          & \textbf{0.5} & -123.1          & 42.4           & \textbf{-33.8}  & -114.5    \\
    $\Delta_{baseline}$&              &              & \cellcolor{red!25}-39.2\%         & \cellcolor{green!25}-10.9\%        & \cellcolor{green!25}\textbf{3.3\%}  & \cellcolor{green!25}-51.01\%  \\
    \bottomrule
    \end{tabular}
    \end{sc}
    \end{small}
    \end{center}
    \vskip -0.1in

    \end{subtable}%
    \begin{subtable}{.5\linewidth}
    \caption{with decision values disabled}
    \label{eval_without_dv}
    \begin{center}
    \begin{small}
    \begin{sc}
    \begin{tabular}{p{0.15cm} p{0.15cm} p{0.15cm} l l l ll}
    \toprule
    $p_{ca}$ & $p_{fc}$ & $p_{rg}$ & $\Sigma r_{ca}$ & $\Sigma r_{fc}$ & $\Sigma r_{rg}$ & $\Sigma\Sigma r_i$ \\
    \midrule
    1            & 1            & 1            & -61.0            & 51.3             & -28.8           & -38.5      \\
    $\Delta_{baseline}$&              &              & ---              & ---              & ---             & ---        \\
    \midrule
    \textbf{1}   & 0            & 0            & \textbf{-77.6}   & 32.0             & -45.0           & -90.6      \\
    $\Delta_{baseline}$&              &              & \cellcolor{red!25}\textbf{-27.2\%} & \cellcolor{green!25}-37.6\%          & \cellcolor{red!25}-56.6\%         & \cellcolor{red!25}-135.26\%  \\
    \midrule
    0            & \textbf{1}   & 0            & -518.2           & \textbf{33.3}    & -58.5           & -543.4     \\
    $\Delta_{baseline}$&              &              & \cellcolor{red!25}-749.1\%         & \cellcolor{red!25}\textbf{-35.0\%} & \cellcolor{red!25}-103.4\%        & \cellcolor{red!25}-1310.52\% \\
    \midrule
    0            & 0            & \textbf{1}   & -126.9           & 31.4             & \textbf{-27.8}  & -123.3     \\
    $\Delta_{baseline}$&              &              & \cellcolor{green!25}108.0\%         & \cellcolor{green!25}-38.7\%          & \cellcolor{green!25}\textbf{3.3\%}  & \cellcolor{green!25}-220.12\%  \\
    \midrule
    \textbf{0.5} & 0.3          & 0.2          & \textbf{-35.7}   & 47.7             & -35.7           & -23.7      \\
    $\Delta_{baseline}$&              &              & \cellcolor{red!25}\textbf{41.6\%}  & \cellcolor{green!25}-7.0\%           & \cellcolor{red!25}-24.1\%         & \cellcolor{red!25}38.54\%    \\
    \midrule
    \textbf{0.5} & 0.2          & 0.3          & \textbf{-40.3}   & 45.2             & -32.6           & -27.7      \\
    $\Delta_{baseline}$&              &              & \cellcolor{green!25}\textbf{34.0\%}  & \cellcolor{green!25}-11.9\%          & \cellcolor{red!25}-13.4\%         & \cellcolor{green!25}28.08\%    \\
    \midrule
    0.2          & \textbf{0.5} & 0.3          & -236.2           & \textbf{49.8}    & -37.8           & -224.2     \\
    $\Delta_{baseline}$&              &              & \cellcolor{red!25}-287.0\%         & \cellcolor{red!25}\textbf{-2.8\%}  & \cellcolor{red!25}-31.5\%         & \cellcolor{red!25}-482.04\%  \\
    \midrule
    0.3          & \textbf{0.5} & 0.2          & -218.9           & \textbf{50.3}    & -38.7           & -207.3     \\
    $\Delta_{baseline}$&              &              & \cellcolor{red!25}-258.7\%         & \cellcolor{red!25}\textbf{-1.8\%}  & \cellcolor{red!25}-34.5\%         & \cellcolor{red!25}-438.10\%  \\
    \midrule
    0.2          & 0.3          & \textbf{0.5} & -86.7            & 41.9             & \textbf{-29.4}  & -74.2      \\
    $\Delta_{baseline}$&              &              & \cellcolor{green!25}-42.1\%          & \cellcolor{red!25}-18.3\%          & \cellcolor{red!25}\textbf{-2.1\%} & \cellcolor{red!25}-92.71\%   \\
    \midrule
    0.3          & 0.2          & \textbf{0.5} & -80.8            & 40.7             & \textbf{-29.1}  & -69.3      \\
    $\Delta_{baseline}$&              &              &\cellcolor{green!25}-32.4\%          & \cellcolor{red!25}-20.7\%          & \cellcolor{red!25}\textbf{-1.3\%} & \cellcolor{red!25}-79.83\%   \\
    \bottomrule
    \end{tabular}
    \end{sc}
    \end{small}
    \end{center}
	\vskip -0.1in
    \end{subtable} 
\end{table*}

\begin{figure}[bt]
\centering
\includegraphics[width=3.5in, trim={2in, 1.5in, 1.5in, 2in}, clip]{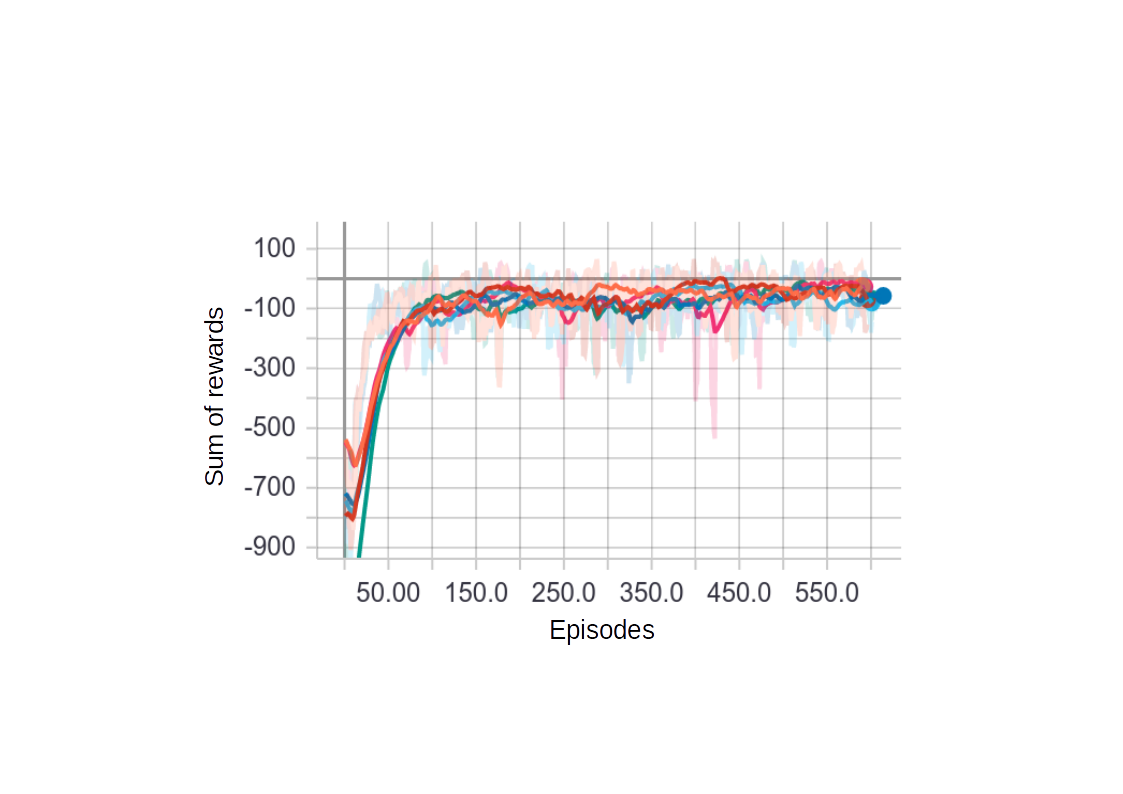}
\caption{The sum of rewards (smoothed) collected by MODQN-DV in cleaner over episodes of training. The plot shows data from 6 different runs.}
\label{sum_of_rewards}
\end{figure}

To evaluate our method we conducted a series of experiments utilizing MODQN-DV and the cleaner environment. In particular we compared the performance of multiple Deep Q-Networks for case a) where decision values were enabled for scalarization and case b) where the decision values were disabled. This comparison gave us a clear indication of the impact of decision values on the performance. We will refer to case (a) as MODQN-DV (b) as to MODQN.

The experiment for both cases (a) and (b) were conducted as follows. First the DQNs were trained using the implementation and parameters as provided in section \ref{implementation} and table \ref{dqn_params}. In case (a) the decision values were trained and used for scalarization. In case (b) decision values were disabled during training and their values were forcefully set to $1$. During training, the user defined priorities for objectives were set to $1$ in all cases (all objectives were weighted equally during scalarization). For each case the training procedure was repeated 6 times and all trained neural networks were saved. As show in figure \ref{sum_of_rewards}, the learning of MODQN-DV is stable over time.

Next, the trained MODQN-DV and MODQN networks were used for evaluation with 10 different sets of user defined priorities $(p_{ca}, p_{fc}, p_{rg})$ as provided in tables \ref{eval_with_dv} and \ref{eval_without_dv}. In a single evaluation, 100 episodes were played. The same sequence of randomly generated map layouts were used for each run. The sum of collected rewards were recorded for each run. For each set of priorities, 6 runs performed by 6 separately trained MODQN-DV and MODQN instances were averaged. 

\subsection{Results}

The results presented in the tables \ref{eval_with_dv} and \ref{eval_without_dv} are averaged sums of collected rewards with respect to each objective, namely: $\Sigma r_{ca}$ for objective (ca) - collision avoidance, $\Sigma r_{fc}$ for objective (fc) - cleaning and $\Sigma r_{rg}$ for objective (rg) - recharging. $\Sigma \Sigma r_i$ is the sum of the sums of rewards - it indicates the total performance of the agent. Priorities $(p_{ca}, p_{fc}, p_{rg})$ correspond to objectives (ca), (fc) and (cg).

The set of priorities: $(p_{ca}=1, p_{fc}=1, p_{rg}=1)$ was used as the baseline for evaluation (also those priories were used during training). For each row in the tables \ref{eval_with_dv} and \ref{eval_without_dv} there is an additional row marked as $\Delta_{baseline}$ with values showing the percentage of gain or loss of collected rewards with respect to the baseline value for each case. The green and red colours of the cells indicate if the reward gain for a particular set of priorities was better compared to the corresponding case in the second table.

The aim of the evaluation was to test how the overall performance of the agent changes when priorities are different from the initial values used during training. As we can see in table \ref{eval_with_dv} on 7 of 9 cases, the use of MODQN-DV helped to preserve (or even increased) the overall performance of the agent in comparison to MODQN. Moreover, in almost all cases, the performance of the agent with respect to the objective with the highest priority (marked in bold in the tables) increased when decision values were used. However, it should be noted that in the baseline case, the overall performance of the agent using decision values was lower compared to the case without decision values. The results show that the proposed solution has a significant impact on the performance when priorities are modified post-training.

It is also worth noticing how the decision values change as the agent moves. As expected, the decision value for a particular objective rises when the agent approach a state where it could receive a reward. For instance, the value of collision avoidance rises significantly when the agent is very close to a wall or an obstacle. Moreover, the decision value drops when the agent is in a state far from receiving a reward. For example, if the agent does not perceive any walls or obstacles, then the collision avoidance decision value is lower than average. The agent thus usually selects the action, which is related to the most promising objective at a particular state.

\section{Conclusions}

In this paper, we presented a method for using multiple Deep Q-Networks to approach multi-objective problems called Multi Objective Deep Q-Networks with Decision Values (MODQN-DV). We introduced decision values to DQNs in order to improve the scalarization of outputs from multiple DQNs. Our method requires only slight modification of existing DQN architectures, while it introduces a number of benefits: 1)~it enables the decomposition of problems in to smaller sub-problems, for which independent DQNs may be trained simultaneously, 2)~it provides a method for robust manipulation of priorities after the training, which also allows to completely disable DQNs responsible for particular behaviour/objective, 3)~it allows to add new objectives to already trained agent without the need of retraining and to tune their impact on the behaviour of the agent.

In the experimental part, we shown that in most cases MODQN-DV improves the performance of the agent, that uses a different set of priorities compared to the training phase. The results are promising, however more tests should be performed using other benchmarks.

In this paper, we also introduce \textit{cleaner} - a benchmark for multi-objective reinforcement learning problems that provides visual state representation. The authors are not aware of any other existing multi-objective benchmark that would be comparable to atari games benchmark or other provided by OpenAI.

In future work we want to improve the performance of MODQN-DV; one possible improvement is the use of common convolutional layers for all DQNs. It is particularly interesting to use MODQN-DV in very complex environments, such as video games. Recently published Starcraft 2 learning environment may be a good choice for further tests of MODQN-DV architecture, as strategy games may be perceived as multi-objective problems.

\bibliography{refs}
\bibliographystyle{icml2018}

\end{document}